\documentclass[11pt,a4paper]{article}
\usepackage[hyperref]{acl2019}
\usepackage{times}
\usepackage{latexsym}

\usepackage{url}
\usepackage{amsmath}
\usepackage{amsfonts}
\usepackage{graphicx}
\usepackage{booktabs}
\usepackage{algorithm}
\usepackage{algorithmic}
\usepackage{multirow}
\usepackage{caption}
\usepackage{subcaption}
\usepackage{tabu}
\usepackage{makecell}
\usepackage{footnote}
\usepackage{footmisc}
\usepackage{color}
\usepackage{scalerel}

\aclfinalcopy 

\title{A Stable Variational Autoencoder for Text Modelling}

\author{Ruizhe Li\textsuperscript{$\spadesuit$}, Xiao Li\textsuperscript{$\spadesuit$}, Chenghua Lin\textsuperscript{$\heartsuit$}, Matthew Collinson\textsuperscript{$\spadesuit$} and Rui Mao\textsuperscript{$\spadesuit$}\\
\textsuperscript{$\spadesuit$}Department of Computing Science, University of Aberdeen, UK\\
\{\texttt{r02rl17, x.li, matthew.collinson, r03rm16}\}\texttt{@abdn.ac.uk}\\
\textsuperscript{$\heartsuit$}Department of Computer Science, University of Sheffield, UK\\
\texttt{c.lin@sheffield.ac.uk}\\
}

\date{}

\begin{document}
\maketitle
\begin{abstract}

Variational Autoencoder (VAE) is a powerful method for learning representations of high-dimensional data. However, VAEs can suffer from an issue known as latent variable collapse (or KL loss vanishing), where the posterior collapses to the prior and the model will ignore the latent codes in generative tasks. Such an issue is particularly prevalent when employing VAE-RNN architectures for text modelling~\cite{bowman2016generating}. In this paper, we present a simple architecture called holistic regularisation VAE (HR-VAE), which can effectively avoid latent variable collapse. 
Compared to existing VAE-RNN architectures, we show that our model can achieve much more stable training process and can generate text with significantly better quality. 
\end{abstract}

\section{Introduction}

Variational Autoencoder (VAE)~\cite{kingma2013auto} is a powerful method for learning representations of high-dimensional data. However, recent attempts of applying VAEs to text modelling are still far less successful compared to its application to image and speech~\cite{bachman2016architecture,fraccaro2016sequential,semeniuta2017hybrid}. When applying VAEs for text modelling, recurrent neural networks (RNNs)\footnote{NB: here we refer RNN to any type of recurrent neural architectures including LSTM and GRU.} are commonly used as the architecture for both encoder and decoder~\cite{bowman2016generating,xu2018spherical,dieng2019avoiding}. While such a VAE-RNN based architecture allows encoding and generating sentences (in the decoding phase)  with variable-length effectively, it is also vulnerable to an issue known as latent variable collapse (or KL loss vanishing), where the posterior collapses to the prior and the model will ignore the latent codes in generative tasks.

\begin{figure*}[tb]
     \centering
     \begin{subfigure}[b]{0.47\textwidth}
         \centering
         \includegraphics[width=\textwidth]{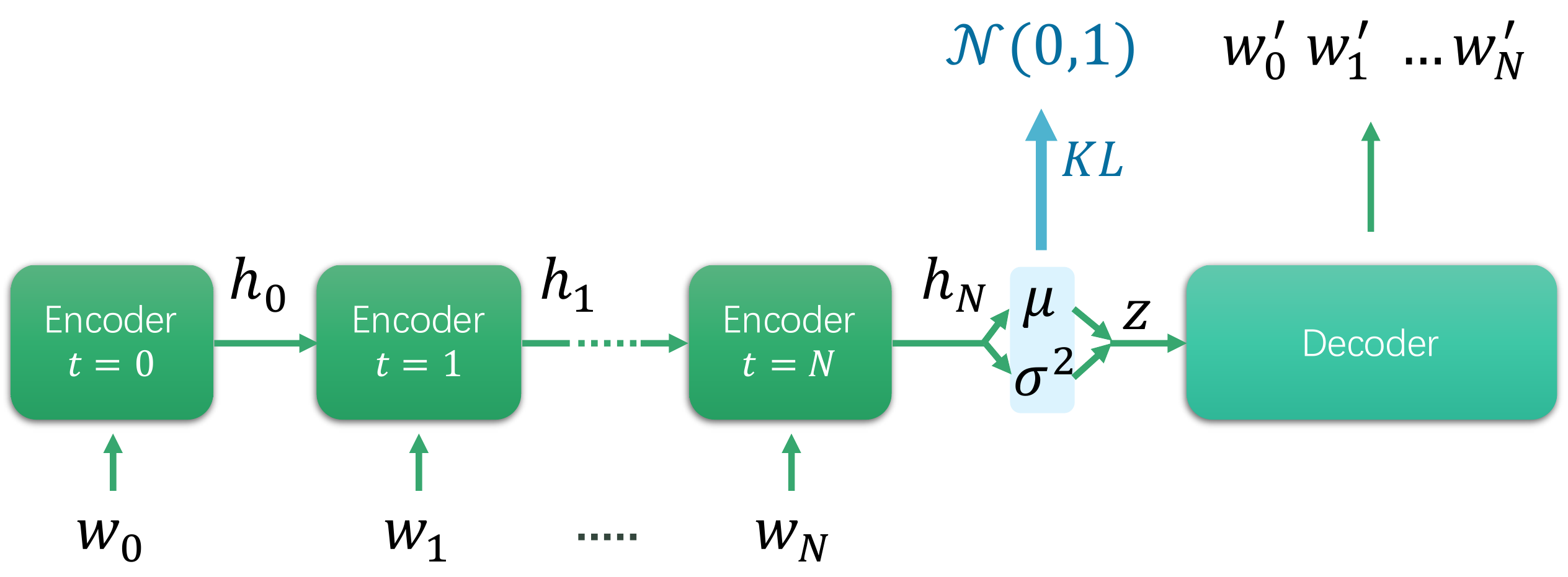}
         \caption{ }
         \label{fg:var-structure}
     \end{subfigure}
     \hfill
     \begin{subfigure}[b]{0.47\textwidth}
         \centering
         \includegraphics[width=\textwidth]{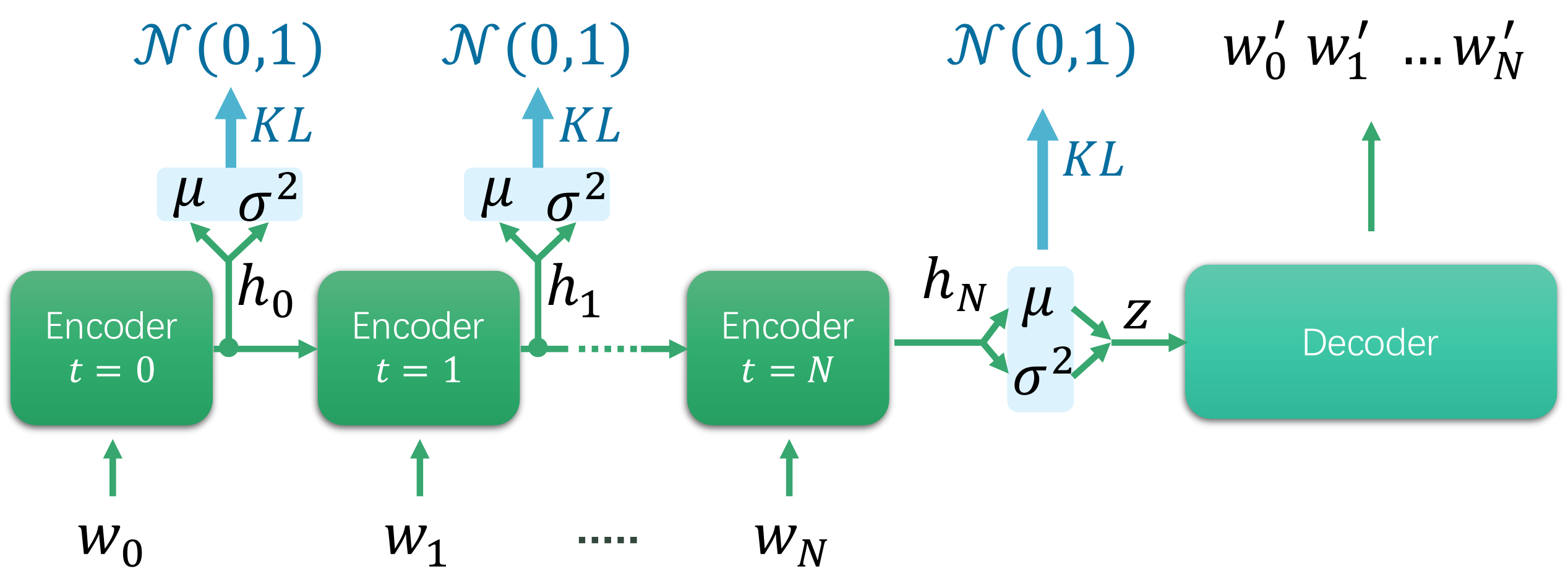}
         \caption{ }
         \label{fg:our-structure}
     \end{subfigure}
        \caption{(a) The typical architecture of RNN-based VAE; (b) the proposed HR-VAE architecture.}
        \label{fig:comparison_VAE}
\end{figure*}

Various efforts have been made to alleviate the latent variable collapse issue. \citet{bowman2016generating} uses  KL annealing, where a variable weight is added to the KL term in the cost function at training time. 
\citet{yang2017improved} discovered that there is a trade-off between the contextual capacity of the decoder and effective use of encoding information, and developed a dilated CNN as decoder which can vary the amount of conditioning context. They also introduced a loss clipping strategy in order to make the model more robust. 
\citet{xu2018spherical} addressed the problem by replacing the standard normal distribution for the prior with the von Mises-Fisher (vMF) distribution. With vMF, the KL loss only depends on the concentration parameter  which is fixed during training and testing, and hence results in a constant KL loss. 
In a more recent work, \citet{dieng2019avoiding} avoided latent variable collapse by including skip connections in the generative model, where the skip connections enforce strong links between the latent variables and the likelihood function.

Although the aforementioned works show effectiveness in addressing the latent variable collapse issue to some extent, they either require carefully engineering to balance the weight between the reconstruction loss and KL loss~\cite{bowman2016generating,sonderby2016train}, or resort to designing more sophisticated model structures~\cite{yang2017improved,xu2018spherical,dieng2019avoiding}.

In this paper, we present a simple architecture  called holistic regularisation VAE (HR-VAE), which can effectively avoid latent variable collapse. In contrast to existing VAE-RNN models for text modelling which merely impose a standard normal distribution prior on the last hidden state of the RNN encoder, our HR-VAE model imposes regularisation for all hidden states of the RNN encoder. Another advantage of our model is that it is generic and can be applied to any existing VAE-RNN-based architectures.

We evaluate our model against several strong baselines which apply VAE for text modelling~\cite{bowman2016generating,yang2017improved,xu2018spherical}. We conducted experiments based on two public benchmark datasets, namely, the Penn Treebank dataset~\cite{marcus1993building} and the end-to-end (E2E) text generation dataset~\cite{novikova2017e2e}. Experimental results show that our HR-VAE model not only can effectively mitigate the latent variable collapse issue with a stable training process, but also can give better predictive performance than the baselines, as evidenced by both quantitative (e.g., negative log likelihood and perplexity) and qualitative evaluation. The code for our model is available online\footnote{\url{https://github.com/ruizheliUOA/HR-VAE}}.

\section{Methodology}

\subsection{Background of VAE}

A variational autoencoder (VAE) is a deep generative model, which combines variational inference with deep learning. The VAE modifies the conventional autoencoder architecture by replacing the deterministic latent representation $\mathbf{z}$ of an input $\mathbf{x}$ with a posterior distribution $P(\mathbf{z}|\mathbf{x})$, and imposing a prior distribution on the posterior, such that the model allows sampling from any point of the latent space and yet able to generate novel and plausible output. 
The prior is typically chosen to be standard normal distributions, i.e.,  $P(\mathbf{z}) = \mathcal{N}(\mathbf{0},\mathbf{1})$, such that the KL divergence between posterior and prior can be computed in closed form~\cite{kingma2013auto}.    

To train a VAE, we need to optimise the marginal likelihood $P_{\theta}(\mathbf{x})=\int{P(\mathbf{z})P_{\theta}(\mathbf{x}|\mathbf{z})d\mathbf{z}}$, where the log likelihood can  take following form:

\begin{equation}
    \log P_{\theta}(\mathbf{x})=\mathcal{L}(\theta,\phi;\mathbf{x}) + \text{KL}\left(Q_{\phi}(\mathbf{z}|\mathbf{x})\|P_{\theta}(\mathbf{z}|\mathbf{x})\right)
\end{equation}
\begin{align} \label{eq:vae_loss}
    \mathcal{L}(\theta,\phi;\mathbf{x})&=\mathbb{E}_{Q_{\phi}(\mathbf{z}|\mathbf{x})}[\log P_{\theta}(\mathbf{x}|\mathbf{z})] \nonumber \\ & \qquad -\text{KL}\left(Q_{\phi}(\mathbf{z}|\mathbf{x})\|P(\mathbf{z})\right)
\end{align}
Here $Q_{\phi}(\mathbf{z}|\mathbf{x})$ is the variational approximation for the true posterior $P_{\theta}(\mathbf{z}|\mathbf{x})$. Specifically,  $Q_{\phi}(\mathbf{z}|\mathbf{x})$ can be regarded as an encoder (a.k.a. the recognition model) and $P_{\theta}(\mathbf{x}|\mathbf{z})$ the decoder (a.k.a. the generative model). Both encoder and decoder are implemented via neural networks. 
As proved in \cite{kingma2013auto}, optimising the marginal log likelihood is essentially equivalent to maximising
$\mathcal{L}(\theta,\phi;\mathbf{x})$,  i.e., the evidence lower bound (ELBO), which consists of two terms. The first term is the expected reconstruction error indicating how well the model can reconstruct data given a latent variable.  The the second term is the KL divergence of the approximate posterior from prior, i.e., a regularisation pushing the learned posterior to be as close to the prior as possible.

\subsection{Variational Autoendoder with Holistic Regularisation}

In this section, we discuss the technical details of the proposed holistic regularisation VAE   (HR-VAE) model, a general architecture which can effectively mitigate the KL vanishing phenomenon.

Our model design is motivated by one noticeable defect shared by the VAE-RNN based models in previous works~\cite{bowman2016generating,yang2017improved,xu2018spherical,dieng2019avoiding}.  That is, all these models, as shown in Figure~\ref{fg:var-structure}, only impose a standard normal distribution prior on the last hidden state of the RNN encoder, which potentially leads to learning a suboptimal representation of the latent variable and results in model vulnerable to KL loss vanishing. Our hypothesis is that to learn a good representation of data and a good generative model, it is crucial to impose the standard normal prior on all the hidden states of the RNN-based encoder (see Figure~\ref{fg:our-structure}), which  allows a better regularisation of the model learning process.

We implement the HR-VAE model using a two-layer LSTM for both the encoder and decoder.  However, one should note that our architecture can be readily applied to other types of RNN such as GRU. 
For each time stamp $t$ (see Figure~\ref{fg:our-structure}), we concatenate the hidden state $\mathbf{h}_t$ and the cell state $\mathbf{c}_t$ of the encoder. The concatenation (i.e.,  $[\mathbf{h}_t;\mathbf{c}_t]$) is then fed into two linear transformation layers for estimating  $\boldsymbol{\mu}_t$ and  $\boldsymbol{\sigma}^2_t$, which are  parameters of a normal distribution corresponding to the concatenation of $\mathbf{h}_t$ and $\mathbf{c}_t$. Let $Q_{\phi_t}(\mathbf{z}_t | \mathbf{x})=\mathcal{N}(\mathbf{z}_t|\boldsymbol{\mu}_t,\boldsymbol{\sigma}^2_t)$, we wish $Q_{\phi_t}(\mathbf{z}_t | \mathbf{x})$ to be close to a prior $P(\mathbf{z}_t)$, which is a standard Gaussian. Finally, the KL divergence between these two multivariate Gaussian distributions (i.e., $Q_{\phi_t}$ and $P(\mathbf{z}_t)$) will contribute to the overall KL loss of the ELBO. By taking the average of the KL loss at each time stamp $t$, the resulting ELBO takes the following form
\begin{align}\label{eq:our-loss}
    \mathcal{L}(\theta,\phi;\mathbf{x}) &= \mathbb{E}_{Q_{\phi}(\mathbf{z}_{\scaleto{N}{2.5pt}}|\mathbf{x})}[\log P_{\theta}(\mathbf{x}|\mathbf{z}_{\scriptscriptstyle N})] \nonumber \\ & \qquad - \frac{1}{N} \sum_{t=0}^{N}\text{KL}(Q_{\phi_t}(\mathbf{z}_t | \mathbf{x}) \| P(\mathbf{z}_t)).
\end{align}

As can be seen in Eq.~\ref{eq:our-loss}, our solution to the KL collapse issue does not require any engineering for balancing the weight between the reconstruction term and KL loss as commonly the case in existing works~\cite{bowman2016generating,sonderby2016train}. The weight between these two terms of our model is simply $1:1$.

\begin{table*}[tb]
  \centering \small
  \begin{tabular}{lccccc}
    \Xhline{2.5\arrayrulewidth}
    Dataset & Training & Development & Testing & Avg. sent. length & Vocab.\\
    \hline
    PTB   & 42,068 & 3,370 & 3,761 & 21.1 & 10K\\
    E2E   & 42,061  & 4,672 & 4,693 & 22.67   & 2.8K\\
    \Xhline{2.5\arrayrulewidth}
  \end{tabular}
  \caption{The statistics of the PTB and E2E datasets.}
  \label{T:datasets}
\end{table*}

\section{Experimental Setup}

\subsection{Datasets}
We evaluate our model on two public datasets, namely, Penn Treebank (PTB)~\cite{marcus1993building} and the end-to-end (E2E) text generation corpus~\cite{novikova2017e2e}, which have been used in a number of previous works for text generation~\cite{bowman2016generating,xu2018spherical,wiseman2018learning,su2018natural}. PTB consists of more than 40,000 sentences from Wall Street Journal articles whereas the E2E dataset contains over 50,000 sentences of restaurant reviews. The statistics of these two datasets are summarised in Table~\ref{T:datasets}.

\begin{table*}[tb]
  \centering
  \small
  \begin{tabular}{c|cccc|cccc}
  \toprule
  \multirow{3}{*}{Model} &
  \multicolumn{4}{c|}{PTB} &
  \multicolumn{4}{c}{E2E}  \\ \cline{2-9}
   &\multicolumn{2}{c}{Standard} & \multicolumn{2}{c|}{Inputless}  &\multicolumn{2}{c}{Standard} & \multicolumn{2}{c}{Inputless}\\
     & NLL & PPL & NLL & PPL & NLL & PPL & NLL & PPL \\
     \midrule
VAE-LSTM-base & {~~101}$^\dag$ (2$^\dag$) & {~~119}$^\dag$ & {~~125}$^\dag$ (15$^\dag$) & {~~380}$^\dag$ & 50 (1.88) & 5.77 & 101 (5.48) & 34.70 \\
VAE-CNN  & 99 (3.1) & 113 & 121 (16.2) & 323 & 41 (3.02) & 4.23 & 82 (5.95) & 17.81 \\
vMF-VAE & {~~96}$^\dag$ (5.7$^\dag$) & {~~98}$^\dag$ & {~~117}$^\dag$ (18.6$^\dag$) & {~~262}$^\dag$ & 34 (7.63) & 3.29 & 61 (19.58) & 8.52 \\
\midrule
HR-VAE (Ours) & \phantom{0}\textbf{79	(10.4)}	&	\textbf{43} & \phantom{0}\textbf{85 (17.32)}	&	\textbf{54} & \textbf{20 (5.37)}  &  \textbf{2.02}  &  \textbf{38 (7.78)}  &  \textbf{3.74} \\
\bottomrule
  \end{tabular}
  \caption{Language modelling results on the PTB and E2E datasets. $^\dag$ indicates the results which are reported from the prior publications. KL loss is shown in the parenthesis.}
  \label{T:NLL_KL}
\end{table*}

\begin{figure*}[tb]
\centering
\includegraphics[scale=0.5]{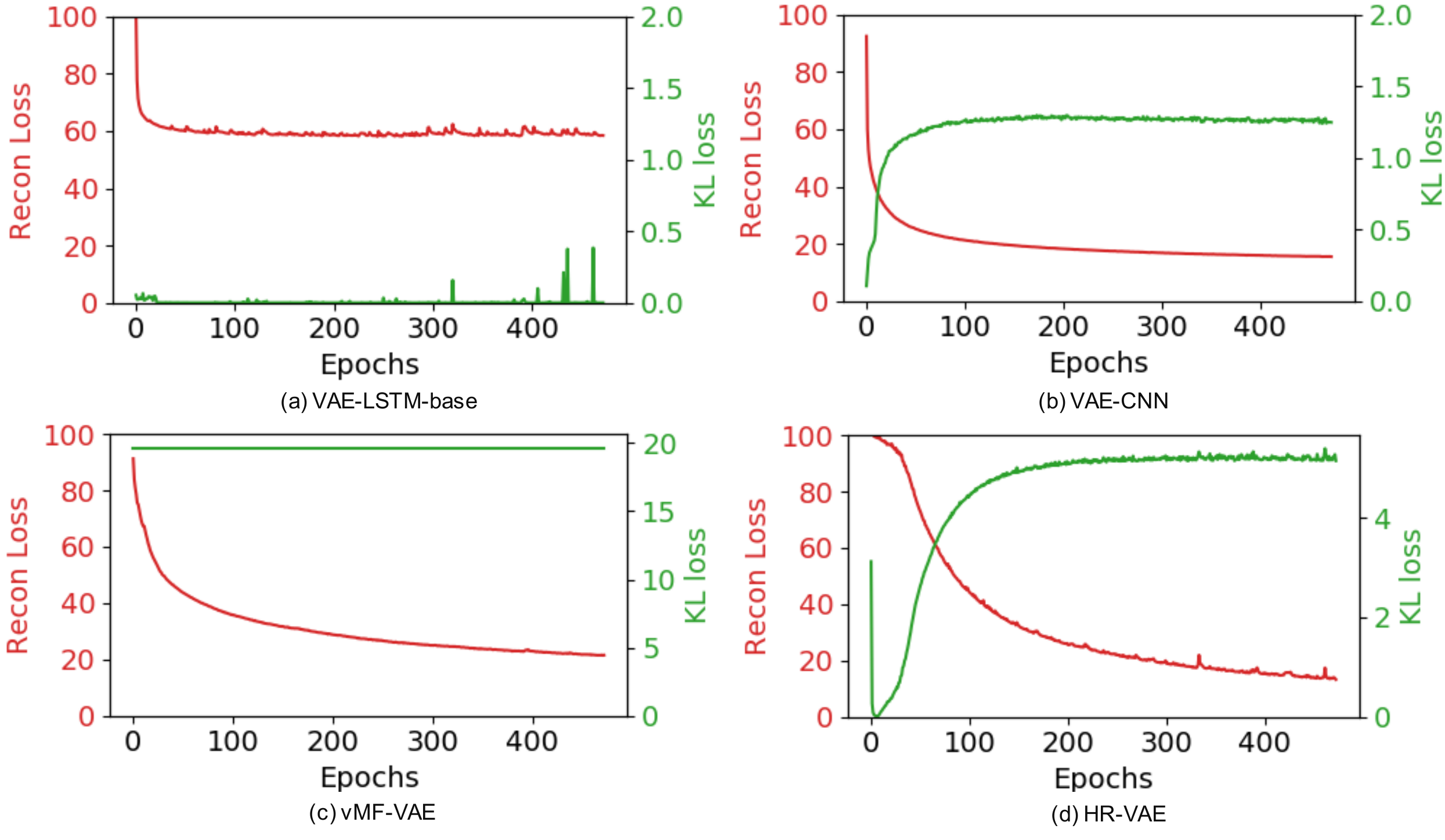}
\caption{Training curves of reconstruction loss and KL loss of (a) VAE-LSTM-base, (b) VAE-CNN, (c) vMF-VAE,  and (d) our model, based on the E2E training set using the inputless setting.}
\label{F: curves}
\end{figure*}

\begin{table*}[tb] 
\small
  \centering 
  \begin{tabular}{c|l} 
    \Xhline{2.5\arrayrulewidth}
    \multirow{5}{*}{\rotatebox[origin=c]{90}{Input}} & 1. blue spice is a coffee shop in city centre . \\
    & 2. giraffe is a coffee shop found near the bakers . \\
    & 3. a pub in the city centre area called blue spice \\
    &4. pub located near caf\'e sicilia called cocum with a high customer rating \\
    & 5. the cricketers is a one star coffee shop near the ranch that is not family friendly .\\ \hline
    \multirow{5}{*}{\rotatebox[origin=c]{90}{vMF-VAE}} & 1. blue spice is a coffee in city centre . it is not , and \\
    & 2. cotto is a coffee shop located near the bakers . . is 5 out of \\
    & 3. a coffee in the city city area is blue spice spice . the is is \\
    & 4. located located near caf\'e rouge , cotto has a high customer rating and a customer \\
    & 5. the cricketers is a low rated coffee shop near the bakers that is a star , is is \\ \hline
    \multirow{5}{*}{\rotatebox[origin=c]{90}{Ours}} & 1. blue spice is a coffee shop in city centre .\\ 
    & 2. giraffe is a coffee shop located near the bakers .\\
    & 3. a restaurant in the city centre called blue spice italian \\
    & 4. located place near caf\'e sicilia called punter has a high customer rating \\
    & 5. the cricketers is a one star coffee shop near ranch ranch that is not family friendly .\\
    \Xhline{2.5\arrayrulewidth}
  \end{tabular}
  \caption{Example input sentences from the E2E test dataset (top); sentences reconstructed by vMF-VAE (middle); sentences reconstructed by our model (bottom).}
  \label{T:sentence}
\end{table*}

\subsection{Implementation Details}
For the PTB dataset, we used the train-test split following~\cite{bowman2016generating,xu2018spherical}. 
For the E2E dataset, we used the train-test  split from the original dataset~\cite{novikova2017e2e} and indexed the  words with a frequency higher than 3. 
We represent input data with 512-dimensional word2vec embeddings~\cite{mikolov2013distributed}. 
We set the dimension of the hidden layers of both encoder and decoder to 256. The Adam optimiser~\cite{kingma2014adam} was used for training with an initial learning rate of 0.0001. Each utterance in a mini-batch was padded to the maximum length for that batch, and the maximum batch-size allowed is 128.

\subsection{Baselines}
We compare our HR-VAE model with three strong baselines using VAE for text modelling:

\noindent \textbf{VAE-LSTM-base}\footnote{\url{https://github.com/timbmg/Sentence-VAE}}: A variational autoencoder model which uses LSTM for both  encoder and decoder. KL annealing is used to tackled the latent variable collapse issue~\cite{bowman2016generating};\\
\noindent \textbf{VAE-CNN}\footnote{\url{https://github.com/kefirski/contiguous-succotash}}: A variational autoencoder model with a LSTM encoder and a dilated CNN decoder~\cite{yang2017improved};\\
\noindent \textbf{vMF-VAE}\footnote{\url{https://github.com/jiacheng-xu/vmf_vae_nlp}}: A variational autoencoder model using LSTM for both encoder and decoder where the prior distribution is the von Mises-Fisher (vMF) distribution rather than  a Gaussian distribution~\cite{xu2018spherical}.

\section{Experimental Results}

We evaluate our HR-VAE model in two experimental settings, following the setup of \cite{bowman2016generating,xu2018spherical}. In the \textit{standard setting}, the input to the decoder at each time stamp is the concatenation of latent variable $\mathbf{z}$ and the ground truth word of the previous time stamp. Under this setting, the decoder will be more powerful because it uses the ground truth word as input,
resulting in little information of the training data captured by latent variable $\mathbf{z}$. 
The \textit{inputless setting}, in contrast, does not use the previous ground truth word as input for the decoder. In other words, the decoder needs to predict the entire sequence with only the help of the given latent variable $\mathbf{z}$. In this way, a high-quality representation abstracting the information of the input sentence is much needed for the decoder, and hence enforcing $\mathbf{z}$ to learn the required information.

\noindent\textbf{Overall performance.}~~~Table~\ref{T:NLL_KL} shows the language modelling results of our approach and the baselines. We report negative log likelihood (NLL), KL loss, and perplexity (PPL) on the test set. As expected, all the models have a higher KL loss  in the inputless setting than the standard setting, as $\mathbf{z}$ is required to encode more information about the input data for reconstruction. In terms of overall performance, our model outperforms all the baselines in both datasets (i.e., PTB and E2E). For instance, when comparing with the strongest baseline  vMF-VAE in the standard setting, our model reduces NLL from 96 to 79 and PPL from 98 to 43 in PTB, respectively. In the inputless setting, our performance gain is even higher, i.e., NLL reduced from 117 to 85 and PPL from 262 to 54. A similar pattern can be observed for the E2E dataset. These observations suggest that our approach can learn a better generative model for data.

\noindent\textbf{Loss analysis.}~~~To conduct a more thorough evaluation, we further investigate model behaviours in terms of both reconstruction loss and  KL loss, as shown in Figure~\ref{F: curves}. These plots were obtained based on the E2E training set using the inputless setting.

We can see that the KL loss of VAE-LSTM-base, which uses Sigmoid annealing~\cite{bowman2016generating}, collapses to zero, leading to a poor generative performance as indicated by the high reconstruction loss. The KL loss for both VAE-CNN and vMF-VAE are nonzero, where the former mitigates the KL collapse issue with a KL loss clipping strategy and the latter by replacing the standard normal distribution for the prior with the vMF distribution (i.e., with the vMF distribution, the KL loss only depends on a fixed concentration parameter, and hence results in a constant KL loss). 
Although both VAE-CNN and vMF-VAE outperform VAE-LSTM-base by a large margin in terms of reconstruction loss as shown in Figure~\ref{F: curves}, one should also notice that these two models actually overfit the training data, as their performance on the test set is much worse (cf. Table~\ref{T:NLL_KL}). 
In contrast to the baselines which mitigate the KL collapse issue by carefully engineering the weight between the reconstruction loss and KL loss or choosing a different choice of prior, we provide a simple and elegant solution through holistic KL regularisation, which can effectively mitigate the KL collapse issue and  achieve a better reconstruction error in both training and testing.

\noindent\textbf{Sentence reconstruction.}~ Lastly, we show some sentence examples reconstructed by vMF-VAE (i.e., the best baseline) and our model in the inputless setting using sentences from the E2E test set as input.  As shown in Table~\ref{T:sentence}, the sentences generated by vMF-VAE contain repeated words in quite a few cases, such as \textit{`city city area'} and \textit{`blue spice spice'}. In addition, vMF-VAE also tends to generate unnecessary or unrelated  words at the end of sentences, making the generated sentences ungrammatical. The sentences reconstructed by our model, in contrast, are more grammatical and more similar to the corresponding ground truth sentences than vMF-VAE.

\section{Conclusion}
In this paper, we present a simple and generic architecture called holistic regularisation VAE (HR-VAE), which can effectively avoid latent variable collapse. In contrast to existing VAE-RNN models which merely impose a standard normal distribution prior on the last hidden state of the RNN encoder, our HR-VAE model imposes regularisation on all the hidden states, allowing a better regularisation of the model learning process. Empirical results show that our model can effectively mitigate the latent variable collapse issue while giving a better predictive performance than the baselines.

\section*{Acknowledgment}
This work is supported by the award made by the UK Engineering and Physical Sciences Research Council (Grant number: EP/P011829/1).

\bibliography{acl2019}
\bibliographystyle{acl_natbib}

\appendix

\end{document}